\documentclass{article}

 \usepackage[preprint]{neurips_2026}

\usepackage[utf8]{inputenc} 
\usepackage[T1]{fontenc}    
\usepackage{hyperref}       
\usepackage{url}            
\usepackage{booktabs}       
\usepackage{amsfonts}       
\usepackage{nicefrac}       
\usepackage{microtype}      
\usepackage{xcolor}         

\usepackage{amsmath}
\usepackage{multirow}
\usepackage{graphicx}
\usepackage{algorithm}
\usepackage{algpseudocode}
\title{TRACER: Training-Free Closed-Loop Structured Inference for Traffic Accident Reconstruction}

\author{%
  Yanchen Guan, Chengyue Wang, Bin Rao, Haicheng Liao,\\
  \textbf{Jiaxun Zhang, Shang Gao, Chengzhong Xu, and Zhenning Li*}
   \AND
  State Key Laboratory of Internet of Things for Smart City\\
  University of Macau\\
  Macau SAR, 999078, China \\
}

\begin{document}

\maketitle

\begin{abstract}

Traffic accident reconstruction is a forensic inverse problem that requires recovering physically consistent motion from sparse and heterogeneous evidence. Existing learning-based approaches predominantly optimize for semantic plausibility or visual realism, rather than quantitative agreement with measurable geometry and dynamics. Here, we present TRACER, a training-free framework that formulates reconstruction as a closed-loop structured inference process. Instead of directly generating dense trajectories, our framework constructs and iteratively refines event-anchored motion hypotheses under geometric, kinematic, and interaction constraints, guided by structured case memory and consistency-driven diagnosis. This design enables incremental, interpretable corrections when evidence is insufficient, making the accident reconstruction process more aligned with the workflow of human experts. Experiments on real-world accident data show that TRACER achieves improved geometric fidelity, velocity consistency, and collision accuracy over both data-driven and physics-based baselines.

\end{abstract}

\section{Introduction}

Accident reconstruction aims to recover the pre-impact motion process that led to an observed crash from partial post-hoc evidence~\cite{davis2003bayesian}. This evidence is often sparse, noisy, and heterogeneous, including textual narratives, vehicle attributes, road geometry, scene sketches, and partial collision cues~\cite{wach2016calculation}. The task therefore requires more than generating plausible trajectories: the recovered motion must remain consistent with the accident description, road topology, vehicle kinematics, multi-agent interaction, and final impact configuration~\cite{gao2020vectornet}.

This setting poses a structured inverse inference problem: several motion histories may explain the same report, especially when the evidence only specifies high-level actions such as turning, braking, yielding, stopping, or running a red light~\cite{arora2021survey}. Under such ambiguity, once the generated dense trajectory becomes inconsistent, the source of the error is difficult to localize, making it unclear which stage of the motion evolution process is responsible for the failure. Consequently, direct dense trajectory generation is not well suited for interpretable diagnosis and localized refinement~\cite{ivanovic2021mats}. In contrast, accident reasoning is naturally organized around event-level stages, including upstream approach, maneuver onset, conflict entry, evasive response, and final collision approach~\cite{menzel2018scenarios}. These stages provide a compact interface between sparse evidence and metric trajectory recovery.

Existing accident reconstruction approaches have used accident reports, scene layouts, or simulators to synthesize accident-related trajectories and visualizations~\cite{guo2024sovar,li2025avd2}. However, many reconstructed results remain vulnerable to structured inconsistencies. A trajectory may match the narrative but violate lane topology, reach the accident point while implying an unrealistic speed profile, or approach the collision from a direction incompatible with the reported impact side. 
These failure modes indicate that the central challenge of accident reconstruction is not merely generating a seemingly plausible trajectory, but rather constructing a structured motion hypothesis that remains diagnosable, localizable, and repairable under incomplete evidence.

We propose TRACER, a closed-loop framework for physically constrained inverse motion reconstruction. TRACER represents each vehicle's pre-impact motion as an event-anchored hypothesis composed of sparse control points, segment relations, speed stages, and collision-approach constraints. A structured case memory provides weak priors over route extent, maneuver style, speed tendency, and approach configuration. Large language models are integrated as structured reasoning components across the inference loop: they propose event-level hypotheses, assess semantic and action-level consistency, and generate localized refinements conditioned on checker diagnostics. Deterministic modules handle dense trajectory realization, numerical consistency checking, and edit acceptance, keeping the final reconstruction auditable and physically constrained.

Our contributions are threefold. First, we introduce an event-anchored motion representation that aligns with the granularity of accident evidence and supports interpretable diagnosis and localized repair. Second, we develop a closed-loop structured consistency minimization framework that couples LLM-assisted planning, hybrid checking, LLM-guided refinement, and deterministic dense realization. Third, we incorporate structured case memory as a weak prior mechanism that regularizes underdetermined reconstruction and provides pattern references. Experiments on real-world accident reconstruction data show that TRACER improves geometric accuracy, speed consistency, and collision-level fidelity over representative trajectory-generation baselines.

\section{Related Work}

Traffic accident reconstruction is a forensic inverse problem that recovers the causal and kinematic process of a crash from heterogeneous evidence, including measurements, sketches, vehicle damage, and witness or driver statements~\cite{zheng2020determinants,struble2020automotive,fernandes2018application,ryan2024accident}. Conventional reconstruction relies on trained investigators to integrate physical evidence into structured reports for legal, engineering, and safety analyses~\cite{rivers2010technical,rivers2006evidence}. Although evidentially rigorous, this process requires substantial expertise, specialized equipment, and prolonged on-site operations~\cite{carper2000forensic,raviv2017analyzing,faizan2021forensic,smith1957physical}, limiting its scalability.

Recent learning-based methods seek to automate reconstruction by using accident reports as semantic inputs~\cite{chen2025transforming,jiao2018virtual,li2024steering}. Existing approaches include language-conditioned simulation, such as SoVAR~\cite{guo2024sovar}; text-conditioned visual generation, such as AVD2~\cite{li2025avd2}; and LLM-assisted simulation pipelines, such as AccidentSim~\cite{zhang2025accidentsim}. These methods improve semantic controllability or visual plausibility, but often leave key physical states, geometric consistency, and quantitative trajectory fidelity insufficiently constrained. Guan et al.~\cite{guan2026learningphysicallygroundedtraffic} move toward evidence-grounded reconstruction by coupling textual cues with scene geometry and motion constraints, yet existing methods still largely rely on one-shot generation without a structured intermediate representation that can be validated and revised.

This limitation is critical under sparse, noisy, and heterogeneous evidence, where multiple motion histories may appear plausible. Human experts typically resolve such ambiguity through iterative hypothesis construction and refinement, reconciling semantic, geometric, and physical constraints. Motivated by this process, we formulate accident reconstruction as structured closed-loop inference with explicit intermediate hypotheses, retrieval-based priors, and consistency-guided refinement.

\section{Methodology}
Given an accident case, our goal is to recover a dense multi-vehicle pre-impact motion history that is consistent with the available semantic, geometric, and collision evidence. We denote the input evidence as
\begin{equation}
    X =
    \{x^{\mathrm{text}}, x^{\mathrm{veh}}, x^{\mathrm{road}}, p^{\star}, q^{\star}, c^{\star}, w\},
\end{equation}
where $x^{\mathrm{text}}$ is the accident summary, $x^{\mathrm{veh}}$ contains vehicle-level attributes such as travel direction, pre-event movement, attempted avoidance maneuver, and speed limit, $x^{\mathrm{road}}$ denotes road and lane geometry, $p^{\star}$ is the accident point, $q^{\star}$ is an impact-side cue, $c^{\star}$ is the collision pair when available, and $w$ denotes weak scene priors. The output is a dense pre-impact trajectory set
\begin{equation}
    \hat{T}
    =
    \{\hat{\mathbf{s}}_v(t_n)\}_{v=1}^{V},
    \qquad
    \hat{\mathbf{s}}_v(t_n)
    =
    [\hat{x}_v(t_n),\hat{y}_v(t_n),\hat{u}_v(t_n)],
\end{equation}
where $V$ is the number of involved vehicles and $t_n$ indexes the pre-impact time steps.

\begin{figure}[t]
\centering 
\includegraphics[width=0.90\textwidth]{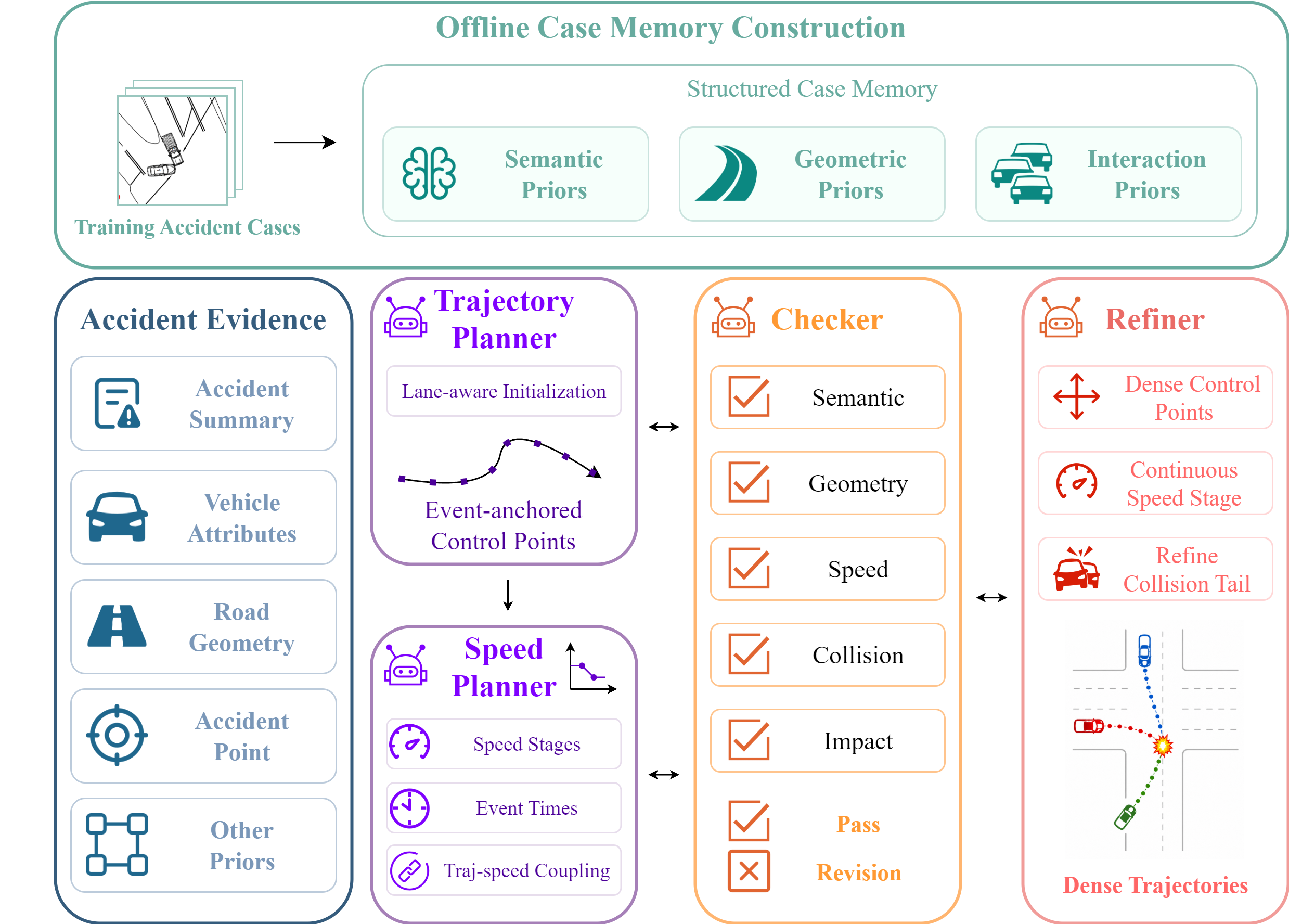}
\caption{Overview of TRACER. Our framework reconstructs pre-impact trajectories through closed-loop constrained inference. Structured memory provides weak priors, the planner generates an event-anchored hypothesis, the checker diagnoses semantic, action-level, geometric, kinematic, and collision inconsistencies, and the refiner repairs the hypothesis with the criticism from checker.}
\label{framework}
\end{figure}

Figure~\ref{framework} illustrates the overall framework. TRACER reconstructs $\hat{T}$ through a structured hypothesis. 
\begin{equation}
    H =
    \{H_v\}_{v=1}^{V}
\end{equation}
denote an event-level trajectory hypothesis, where each $H_v$ contains sparse control points, segment relations, and speed-temporal variables. The method seeks a hypothesis that minimizes a structured consistency energy:
\begin{equation}
\begin{aligned}
    \mathcal{E}(H;X,M)
    =
    &\lambda_{\mathrm{sem}}\mathcal{E}_{\mathrm{sem}}
    + \lambda_{\mathrm{act}}\mathcal{E}_{\mathrm{act}}
    + \lambda_{\mathrm{geo}}\mathcal{E}_{\mathrm{geo}} 
    + \lambda_{\mathrm{spd}}\mathcal{E}_{\mathrm{spd}}
    + \lambda_{\mathrm{kin}}\mathcal{E}_{\mathrm{kin}}
    + \lambda_{\mathrm{col}}\mathcal{E}_{\mathrm{col}} ,
\end{aligned}
\end{equation}
where $M=M(X)$ denotes retrieved case memory. The terms respectively measure consistency with the accident description, action sequence, road geometry, path--speed relation, vehicle kinematics, and collision configuration. Since this objective involves discrete semantic judgments, geometric constraints, and collision-specific reasoning, TRACER approximates its minimization through a closed-loop constrained inference process:
\begin{equation}
    X \rightarrow M(X) \rightarrow H^{(0)}
    \rightarrow C^{(0)} \rightarrow H^{(1)}
    \rightarrow \cdots \rightarrow \hat{T},
\end{equation}
where the planner proposes an initial event-level hypothesis, the checker diagnoses violations of the consistency energy, the refiner reduces the diagnosed inconsistencies, and the realizer converts the refined hypothesis into dense trajectories. 

\subsection{Structured Case Memory}

Sparse accident evidence often leaves multiple motion histories plausible. To reduce this ambiguity, we use a structured case memory from the training split. Each memory item is represented as
\begin{equation}
    m_i =
    \{z_i^{\mathrm{sem}}, z_i^{\mathrm{geo}}, z_i^{\mathrm{int}}, \pi_i\},
\end{equation}
where $z_i^{\mathrm{sem}}$ summarizes semantic cues such as vehicle count, pre-event movement, avoidance maneuver, impact-side evidence, and speed-related information; $z_i^{\mathrm{geo}}$ summarizes road layout and lane-structure cues; $z_i^{\mathrm{int}}$ describes interaction and approach relations; and $\pi_i$ stores compact motion priors, including approximate path extent, approach angle, dominant motion pattern, speed tendency, and braking style.

The retrieved memory set is
\begin{equation}
    M(X)=\{m_{(1)},\dots,m_{(K)}\},
\end{equation}

Each training case is converted into a memory descriptor containing semantic, interaction, road-geometric, speed, and trajectory summaries. The retrieval first ranks cases by semantic and interaction compatibility, and then reranks the top candidates by road-geometry similarity. The resulting top-$K$ cases are used as structured references for the planner and speed planner.
The memory is used as a weak prior over the reconstruction hypothesis, because textual evidence is often insufficient to determine metric motion uniquely, while similar cases can still provide useful regularities about how vehicles approach conflict regions under comparable road and interaction conditions. It narrows the plausible range of approach paths, maneuver styles, speed profiles, and collision-approach configurations, therefore regularizes the hypothesis space. The final reconstruction remains determined by the current case evidence and by the subsequent consistency checking and refinement process.

\subsection{Event-Anchored Planner}

\begin{figure}[t]
\centering 
\includegraphics[width=0.90\textwidth]{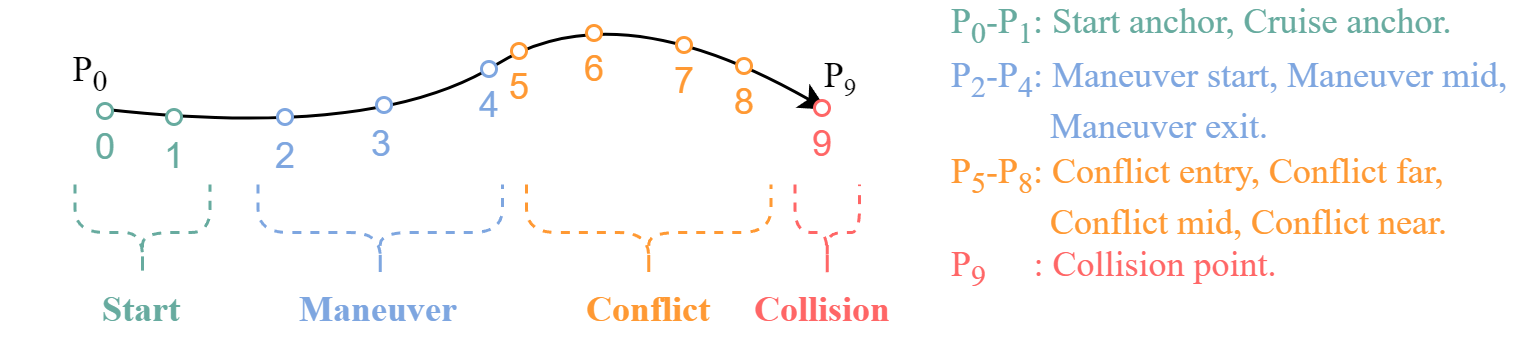}
\caption{Event-anchored trajectory hypothesis. The planner represents each vehicle trajectory using sparse control points aligned with accident-relevant stages, such as upstream approach, maneuver onset, conflict entry, evasive response, and final collision approach. This categorization aligns with the inputs from the original accident reports, which delineate four primary phases: Initial, Motion, Reaction, and Collision.}
\label{anchors}
\end{figure}

The planner constructs an initial structured hypothesis $H^{(0)}$ from the accident evidence and retrieved memory priors. For each vehicle $v$, the hypothesis is represented as
\begin{equation}
    H_v =
    (P_v,R_v,U_v),
\end{equation}
where $P_v$ contains event-anchored control points, $R_v$ contains segment-level motion relations, and $U_v$ contains the speed profile and temporal allocation associated with the event anchors.
\begin{equation}
    P_v=\{\mathbf{p}_{v,k}\}_{k=0}^{K_a-1}.
\end{equation}
In our implementation, $K_a=10$, corresponding to a compact accident-oriented motion skeleton, as shown in Figure~\ref{anchors}.

 The event-anchored representation is more suitable for accident reconstruction than one-shot dense trajectory generation. Accident reports usually describe motion through a small number of semantically meaningful stages, such as pre-event stages and avoidance stages, rather than through frame-level description. A dense sequence generated directly from such evidence is difficult to verify because an error may be distributed across many time steps and to identify the specific action corresponding to each time step. By contrast, an event-level hypothesis exposes the key reconstruction decisions: where the vehicle approaches from, when it enters the conflict region, how it maneuvers or responds, and how it reaches the collision point. These decisions can be checked against road geometry, action semantics, speed feasibility, and impact configuration before dense trajectories are realized. Moreover, when reconstruction errors occur, the event-anchor representation enables the inconsistency to be localized to specific anchors. This establishes an explicit correspondence between the reconstructed trajectory and the input behavioral evidence, making the underlying failure easier to diagnose and supporting localized trajectory editing and refinement.

The planner is LLM-assisted but constrained by the structured hypothesis format. It receives the accident description, vehicle attributes, road geometry, collision cues, and retrieved memory summaries, and proposes event anchors, segment relations, and speed-temporal variables:
\begin{equation}
    H^{(0)}
    =
    f_{\mathrm{plan}}(X,M).
\end{equation}
The geometric part of the hypothesis specifies where the vehicle approaches from, how it traverses the road structure, and how it reaches the collision region. The speed-temporal part specifies a staged pre-impact motion profile, including approximate cruising, maneuver, avoidance, and collision-approach phases. Formally, we write
\begin{equation}
    U_v =
    \left(
    \{u_v(t_n)\}_{n=0}^{T-1},
    \ell_v,
    \tau_v
    \right),
\end{equation}
where $u_v(t_n)$ is the speed at time $t_n$, $\ell_v$ denotes speed-stage labels, and $\tau_v$ denotes event-time allocation for key anchors or motion phases.

The memory priors influence both spatial and temporal components of the initial hypothesis. They suggest plausible route extent, approach angle, maneuver style, speed tendency, and braking behavior, but they do not determine coordinates or dense trajectories directly. This is important because a plausible geometric path may still imply an invalid reconstruction if its length is inconsistent with the inferred speed profile. The planner therefore initializes $P_v$, $R_v$, and $U_v$ jointly, allowing the checker to evaluate semantic, geometric, and path--speed consistency within the same structured hypothesis.

\subsection{Consistency Checker}

The checker evaluates the current hypothesis $H^{(r)}$ by identifying violations of the structured consistency energy. It produces a diagnostic object
\begin{equation}
    C^{(r)}
    =
    f_{\mathrm{check}}(H^{(r)},X,M),
\end{equation}
where $C^{(r)}$ contains violation scores and localized issue descriptions. The diagnosis specifies which vehicle, segment, event anchor, or collision relation contributes to the inconsistency.

The checker combines LLM-based semantic judgment with deterministic physical and geometric evaluation. The semantic consistency term evaluates whether the hypothesis agrees with the accident report and vehicle attributes:
\begin{equation}
    \mathcal{E}_{\mathrm{sem}}
    =
    \sum_{v=1}^{V}
    d_{\mathrm{sem}}(H_v,x^{\mathrm{text}},x^{\mathrm{veh}}).
\end{equation}
The action-consistency term evaluates whether the inferred sequence of actions is compatible with the reported maneuver, avoidance behavior, and interaction process:
\begin{equation}
    \mathcal{E}_{\mathrm{act}}
    =
    \sum_{v=1}^{V}
    d_{\mathrm{act}}(H_v,x^{\mathrm{text}},x^{\mathrm{veh}}).
\end{equation}
These two terms are assessed by the LLM checker because they require interpretation of high-level descriptions such as turning, stopping, swerving, yielding, or running a red light.

The remaining terms are evaluated through explicit computation whenever possible. The geometric term checks lane adherence, travel direction, and road compatibility:
\begin{equation}
    \mathcal{E}_{\mathrm{geo}}
    =
    \sum_{v=1}^{V}
    d_{\mathrm{geo}}(H_v,x^{\mathrm{road}}).
\end{equation}
The path--speed term couples the geometric path length with the speed-implied travel distance. Let $L_v(H)$ be the path length induced by the event anchors and segment relations, and let
\begin{equation}
    D_v(U_v)=\Delta t\sum_{n}u_v(t_n)
\end{equation}
be the distance implied by the speed profile. We use
\begin{equation}
    \mathcal{E}_{\mathrm{spd}}
    =
    \sum_{v=1}^{V}
    \left|
    \frac{L_v(H)}{D_v(U_v)+\epsilon}-1
    \right|.
\end{equation}
The kinematic term penalizes unstable speed variation:
\begin{equation}
    \mathcal{E}_{\mathrm{kin}}
    =
    \sum_{v=1}^{V}
    \frac{
    \max\left(0,\max_n |u_v(t_{n+1})-u_v(t_n)|-3.0\right)
    }{8.0}.
\end{equation}
Finally, the collision term evaluates whether the final approach is compatible with the accident point and available collision cues:
\begin{equation}
    \mathcal{E}_{\mathrm{col}}
    =
    \sum_{(i,j)\in c^{\star}}
    d_{\mathrm{col}}
    (H_i,H_j,p^{\star},q^{\star}).
\end{equation}

The checker does not only score the hypothesis. Its main role is to identify where and why the reconstruction is inconsistent. This localized diagnosis is essential because most reconstruction failures are not global. A trajectory may have a plausible upstream approach but an invalid collision tail, or a correct endpoint but an implausible speed--path relation. By exposing these failure modes at the event and segment level, the checker provides the information needed for targeted refinement.

\subsection{Consistency-Guided Refiner}

The refiner updates the current hypothesis according to the checker diagnosis:
\begin{equation}
    H^{(r+1)}
    =
    f_{\mathrm{refine}}(H^{(r)},C^{(r)},X,M).
\end{equation}
Its objective is to reduce the structured consistency energy:
\begin{equation}
    \mathcal{E}(H^{(r+1)};X,M)
    <
    \mathcal{E}(H^{(r)};X,M).
\end{equation}
The refinement step is local. Rather than regenerating a complete dense trajectory, the refiner edits the event-level hypothesis by adjusting control points, segment relations, speed-stage variables, or event timing. The possible corrections include modifying an upstream approach, changing a maneuver segment, improving path--speed compatibility, or repairing the final collision-approach region. The detailed local refinement procedure is provided in Appendix~\ref{app:local_refinement}.

The refiner is also LLM-assisted. The LLM receives the current structured hypothesis, the checker diagnosis, the accident evidence, and memory priors, and proposes corrections in the same hypothesis space used by the planner. This keeps planning and refinement aligned: both operate on event anchors, segment relations, and speed-temporal variables. Deterministic checks are then used to evaluate whether the proposed update reduces the relevant inconsistency.

This closed-loop design gives TRACER an explicit constrained inference mechanism. The planner produces an initial explanation of the accident process, the checker turns the consistency energy into localized violations, and the refiner modifies the hypothesis to reduce these violations. The loop continues until the hypothesis becomes stable or the predefined refinement budget is reached.

\subsection{Trajectory Realization}

The final stage converts the refined event-level hypothesis into dense trajectories. For each vehicle, the event anchors and segment relations define a continuous geometric path
\begin{equation}
    \Gamma_v(s),
    \qquad
    s\in[0,L_v],
\end{equation}
where $s$ is arc length and $L_v$ is the total path length. This construction separates spatial path generation from temporal parameterization. The anchors and segment relations determine the road-consistent path, while the speed and temporal variables determine how the vehicle progresses along that path during the pre-impact window.

Given the speed profile $u_v(t_n)$, we set $s_v(t_0)=0$ and update the traveled distance as
\begin{equation}
    s_v(t_{n+1})
    =
    \operatorname{clip}
    \left(
    s_v(t_n)+\Delta t\,u_v(t_n),
    0,
    L_v
    \right).
\end{equation}
The dense trajectory is then obtained by evaluating the path at the corresponding arc-length positions:
\begin{equation}
    [\hat{x}_v(t_n),\hat{y}_v(t_n)]
    =
    \Gamma_v(s_v(t_n)),
    \qquad
    \hat{u}_v(t_n)=u_v(t_n).
\end{equation}

For vehicles involved in the collision, the terminal segment is constrained by the accident point and available impact-side evidence. This ensures that the reconstructed trajectories are not only endpoint-aligned, but also consistent with the local contact geometry near impact. The final output is therefore a dense trajectory sequence derived from a memory-regularized, checked, and refined event-level reconstruction hypothesis.

\section{Experiments}

\subsection{Dataset}

We evaluate our method on CISS-REC, a real-world traffic accident reconstruction dataset. CISS-REC contains 6,217 in-depth accident cases collected in the United States, with 5,217 cases used for training and 1,000 held out for testing. Each case provides heterogeneous reconstruction evidence, including textual accident reports, total-station-based scene measurements, high-precision maps, vehicle trajectories, and EDR signals.

As our method is training-free, the original training split is not used for parameter optimization. Instead, it is converted into a structured case memory that provides retrieval-based accident priors. All methods are evaluated on the same 1,000-case test split to ensure a fair comparison with learning-based baselines.

\subsection{Evaluation metrics}

We assess reconstruction quality from geometric, kinematic, collision, semantic, and physical perspectives. Geometric fidelity is measured by average keypoint distance (AKD), which computes the mean Euclidean error between reconstructed and reference vehicle positions, and by average accident-point distance (AAPD), which further decomposes collision localization error into tangential and normal components, denoted as $AAPD_{\mathrm{tan}}$ and $AAPD_{\mathrm{norm}}$. Kinematic accuracy is evaluated by average velocity deviation (AVD) against EDR-derived reference speeds.

To evaluate event-level collision consistency, collision rate (CR) measures whether the reconstructed trajectories produce a valid contact event, while collision surface accuracy (CSA) measures whether the reconstructed impact region matches the reported contact surfaces. We further report behavior accuracy (BA) and role accuracy (RA) evaluated by LLM to quantify consistency with annotated participant actions and causal roles. Finally, physical motion consistency is evaluated using average acceleration error (AccE) and curvature error (KE), which compare reconstructed vehicle dynamics with the corresponding reference trajectories.

\subsection{Evaluation results}

\begin{table*}[t]
\centering
\caption{Quantitative comparison of different models for accident reconstruction. Distance-based metrics are reported in $m$, velocity-based metrics in $m/s$, and accuracy-based metrics in \%. }
\resizebox{0.98\linewidth}{!}{
\begin{tabular}{c c c cc c c c c c c c}
\hline
\multirow{2}{*}{Model} & \multirow{2}{*}{AKD↓} & \multirow{2}{*}{AVD} & \multicolumn{2}{c}{AAPD} & \multirow{2}{*}{CR} & \multirow{2}{*}{CSA} & \multirow{2}{*}{BA} & \multirow{2}{*}{RA} & \multirow{2}{*}{AccE} & \multirow{2}{*}{KE} \\
\cline{4-5}
& & & tan & norm & & & & \\
\hline

STGCN~\cite{yu2017spatio}& 12.15 & 4.82 & 9.09 & 3.71 & 19.70 & 28.79 & 4.95 & 74.73 & 25.35 & 1.03\\
Spline~\cite{schumaker2007spline}& 12.01 & 7.42 & 16.57 & 6.21 & 22.73 & 53.73 & 3.30 & 61.54 &14.44  &0.18\\
 LSTM~\cite{van2020review} & 11.88 & 10.85 &9.28  &4.68  &16.67  &41.79  & 40.11 & 76.81 & 4.10 &0.43\\
  Wayformer~\cite{nayakanti2022wayformer}& 11.64 & 5.72 & 8.83 & 3.52 & 22.73 & 32.58 &5.49  & 75.27 &39.32  &1.08\\
 Spline with temporal allocation~\cite{guan2026learningphysicallygroundedtraffic} & 11.35 & 6.62 & 8.06 & 3.45 & 21.21 & 44.78 &2.20  & 59.34 & 11.71 &0.20\\
HiVT~\cite{zhou2022hivt}& 11.34 & 6.12 & 8.54 & 3.56 & 19.70 & 41.04 & 1.65 & 72.53 & 51.65 &1.05\\
Kinematic fitting~\cite{guan2026learningphysicallygroundedtraffic} & 7.80 & 8.90 & 9.49 & 5.41 & 22.73 & 20.45 &42.86  & 79.01 & 3.67 & 0.02\\
 VectorNet~\cite{gao2020vectornet} & 7.63 & 6.36 & 6.37 & 3.82 & 40.91 & 26.87 &40.11  & 87.91 & 33.40 & 0.89\\
PC-Crash~\cite{steffan1996collision} & 6.29 & 9.53 & 5.31 & 3.29 & 98.48 & 8.96 &37.91  & 84.07 & 21.81 &0.14\\
Momentum-Energy~\cite{zhang2006virtual} & 6.09 & 13.81 & 5.63 & 3.43 & 100.00& 17.91 &17.58  & 73.08 & 28.46 &0.05\\
Physically Grounded TAR~\cite{guan2026learningphysicallygroundedtraffic} & 4.98 & 6.84 & 3.47 & 1.41 &85.71 & 56.35 & 59.69 & 87.88 & 2.23 & 0.04\\

 \hline 
 TRACER & 4.54 & 4.92 & 2.98 & 1.05 &88.27 & 57.06 & 65.66 & 93.25 & 1.87 & 0.05 \\
\hline 
\end{tabular}
}
\label{results}
\end{table*}

Table~\ref{results} presents the performance comparison between the proposed framework and other baseline models on the CISS-REC dataset. All learning-based baselines are adapted to the same reconstruction input contract, including encoded textual evidence, vehicle attributes, road geometry, initial lane, accident-point information, and vehicle- or collision-related conditions. In the experiments, our framework adopts Qwen-3.5-Flash as the default backbone LLM, with the number of retrieved cases set to Top-$K=5$ and the maximum number of refinement iterations set to $3$.

According to the results, the proposed framework achieves the best overall performance in trajectory shape, velocity consistency, collision localization, collision pattern reconstruction, and semantic behavior-role consistency. Notably, although PC-Crash and Momentum-Energy explicitly constrain vehicles to collide at a common point and therefore obtain high collision rates, their simplified collision assumptions lead to poorer collision localization and impact-pattern accuracy.

Compared with the second-best model, which is the first trajectory reconstruction approach designed for weakly supervised settings with physical constraints, our method not only outperforms it across all metrics, but also demonstrates a significant advantage in AVD, achieving performance comparable to STGCN, which primarily captures velocity trends. This improvement can be attributed to two key factors: first, the use of case memory enables the model to learn velocity evolution patterns from similar accident cases; second, the explicit coupling between trajectory geometry and speed planning allows motion patterns to guide velocity hypothesis construction, thereby alleviating the optimization conflict between trajectory and velocity learning. Furthermore, significant improvements in acceleration error also substantiate this view.

\begin{figure}[t]
\centering 
\includegraphics[width=0.90\textwidth]{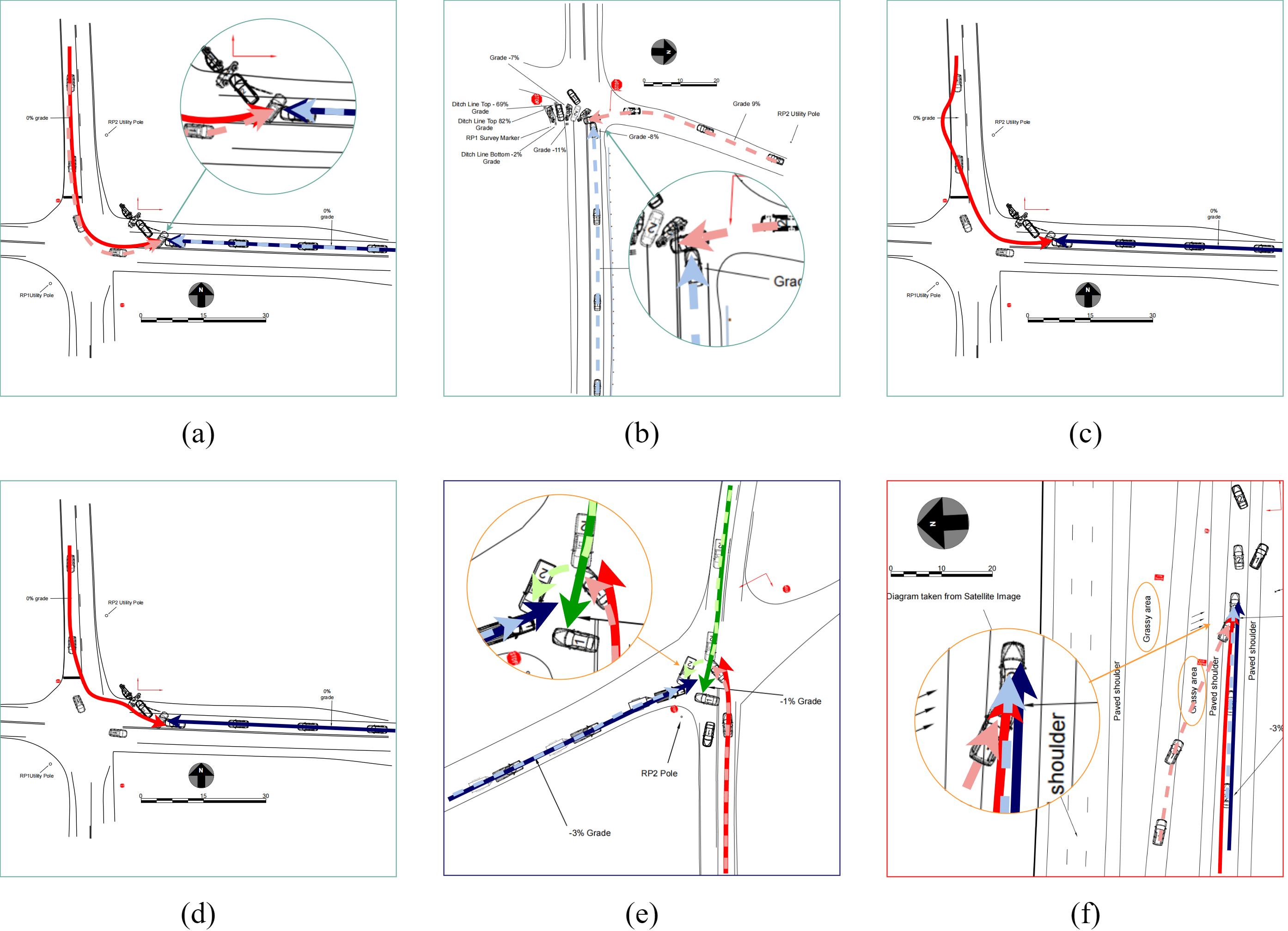}
\caption{Qualitative visualization of accident trajectory reconstruction results. Dark solid trajectories denote the reconstructed motions, while light dashed trajectories denote the original trajectories. 
(a) Original trajectory and reconstructed result of a representative accident case. 
(b) Case memory used for the reconstruction in (a). 
(c) Reconstruction result without the refiner. 
(d) Reconstruction result without case memory. 
(e) Original trajectory and reconstructed result in a multi-vehicle accident scenario. 
(f) Original trajectory and reconstructed result of a failure case.}
\label{cases}
\end{figure}

Figure~\ref{cases} presents visualization results of several representative reconstruction cases. Figure~\ref{cases}(a) shows a turning-versus-straight conflict scenario, while Fig.~\ref{cases}(b) illustrates the corresponding retrieved case memory selected based on behavioral and contextual conditions. The motion patterns in the two cases are similar, yet they are not entirely identical in terms of scale, direction, and specific details. The final result does not directly replicate the retrieved case; instead, it utilizes it as a weak prior and re-infers the outcome based on current evidence. Figures~\ref{cases}(c) and~\ref{cases}(d) present reconstruction results after ablating the refiner and case memory modules, respectively. It can be observed that for relatively complex motion patterns, such as turning vehicles, the combination of case memory and the refiner effectively corrects implausible trajectory segments through weak prior guidance and consistency-based repair. Moreover, due to the event-anchored representation, the refiner is able to adjust trajectory segments within the maneuver phase while preserving the consistency of the conflict phase, enabling local trajectory correction without requiring full regeneration.

Figures~\ref{cases}(e) and~\ref{cases}(f) further highlight several limitations of the current framework. Figure~\ref{cases}(e) depicts a multi-vehicle scenario. Although the reconstructed trajectories successfully reproduce the collision events described in the report, the model fails to capture abrupt motion changes between multiple collisions of the same vehicle, instead producing overly smooth trajectories. Figure~\ref{cases}(f) presents a failure case, where the red vehicle initially travels in an open area between two grass regions. As a result, the boundaries on both sides of its initial position are not recognized as valid road edges, causing the vehicle location to fall outside the candidate lane set. Consequently, the model assigns this vehicle to the same lane as the blue vehicle based on semantic cues, leading to a significant deviation in the initial position and ultimately affecting the overall reconstruction outcome.

\subsection{Ablation studies}

Table~\ref{Ablation} shows that the proposed components contribute to different aspects of reconstruction. Removing case memory produces the overall degradation, which suggests that memory is important for resolving the underdetermined global approach geometry and motion scale. Using only the top-$3$ retrieved cases recovers most of the performance, but remains slightly worse than the full model, indicating that a richer memory context provides more stable priors.

Speed--path coupling mainly affects velocity consistency. Without it, AVD increases to $6.37$, whereas AKD, CR, and CSA remain close to the full model. This indicates that the coupling term primarily aligns path length with speed evolution rather than changing the collision outcome. In contrast, removing the checker and refiner mainly harms local accident-point alignment, which confirms the role of closed-loop diagnosis and refinement in correcting local geometric inconsistencies. The event-anchor representation provides substantial improvements in spatial localization, collision consistency, and collision pattern reconstruction for traffic accident reconstruction. By generating and evaluating trajectories according to accident-oriented stages, the framework achieves stronger interpretability, improved localized reparability, and more stable collision consistency.

The refinement-round ablation shows that one round already achieves performance close to the full model, and two rounds provide only marginal additional gain. Removing collision-tail repair has a focused effect on terminal collision quality: CR drops from $88.27\%$ to $82.31\%$, CSA drops from $57.06\%$ to $50.00\%$, and both AAPD errors increase. This demonstrates that terminal repair improves impact-point and contact-side fidelity while leaving the global path largely unchanged. Overall, the results indicate that memory, speed--path coupling, checker-refiner inference, and collision-tail repair address complementary reconstruction errors.

\begin{table*}[t]
\centering
\caption{Ablation study of key modules of our proposed model.}
 \resizebox{0.60\linewidth}{!}{
\begin{tabular}{c c c cc c c }
\hline \specialrule{0em}{1pt}{1pt}
\multirow{2}{*}{Model} & \multirow{2}{*}{AKD} & \multirow{2}{*}{AVD} & \multicolumn{2}{c}{AAPD} & \multirow{2}{*}{CR} & \multirow{2}{*}{CSA} \\
\cline{4-5}
& & & tan & norm & & \\
\hline
w/o Case Memory & 6.09 & 6.91 & 3.25  & 1.31 & 86.27 & 46.08 \\
Memory Top K = 3  & 4.61 & 5.46 & 3.12  & 1.19 & 86.76 & 55.09 \\
 w/o Speed-path Coupling & 4.74 & 6.37 & 3.05  & 1.06 & 88.27 & 57.06 \\
w/o Checker \& Refiner & 4.84 & 5.09 & 3.82  & 1.37 & 86.23 & 53.92 \\
Max Refine Round = 1   & 4.59 & 4.92 &  2.99 & 1.05 & 88.27 & 57.06 \\
Max Refine Round = 2   & 4.55 & 4.93 & 2.98  & 1.05 & 88.27 & 57.06 \\
w/o Collision Tail Repair & 4.62 & 4.92 &  3.44 & 1.13 & 82.31 & 50.00 \\ 
w/o Event-anchor & 6.47 & 5.59 &  3.53 & 1.66 & 75.71 & 42.72 \\
\hline
FULL & 4.54 & 4.92 & 2.98 & 1.05 &88.27 & 57.06  \\ \hline
\end{tabular}
}

\label{Ablation}
\end{table*}

\section{Conclusion}

We presented TRACER, a training-free framework for traffic accident reconstruction from sparse and heterogeneous evidence. TRACER reconstructs pre-impact motion through event-anchored hypotheses that can be checked and repaired at semantically meaningful stages. Experiments on CISS-REC show that this structured inference design improves geometric fidelity, velocity consistency, and collision-level accuracy. The ablation studies further indicate that structured case memory mainly helps resolve ambiguous approach geometry and motion scale, speed--path coupling improves velocity consistency, and the checker--refiner loop together with collision-tail repair improves local accident-point and contact-side fidelity. TRACER is intended as an assistive reconstruction tool, not an autonomous liability-assessment system. Future work should extend the fixed skeleton to adaptive event graphs, incorporate stronger dynamic-feasibility constraints, and quantify uncertainty under incomplete or ambiguous evidence. In forensic, legal, or insurance-related applications, the reconstructions should therefore be interpreted together with physical evidence and expert judgment.

\newpage
\bibliographystyle{unsrt}
\bibliography{refs}

\newpage

\newpage

\end{document}